\documentclass[runningheads]{llncs}
\usepackage[T1]{fontenc}
\usepackage{cite}
\usepackage{amsmath,amssymb,amsfonts}
\usepackage{algorithmic}
\usepackage{graphicx}
\usepackage{textcomp}
\usepackage{xcolor}

\usepackage{algorithm}
\usepackage{comment}
\usepackage{enumitem}
\usepackage{multirow}
\usepackage{booktabs}
\usepackage{graphicx}
\usepackage{colortbl}
\usepackage{subcaption}
\usepackage{hyperref}

\definecolor{LightPink}{rgb}{1.0, 0.8, 0.8} 
\definecolor{Peach}{rgb}{1.0, 0.9, 0.8}

\begin{document}
\title{ MCIGLE: Multimodal Exemplar-Free Class-Incremental Graph Learning }
%
%

\author{
Haochen You\inst{1}\orcidID{0009-0008-9178-2912}\thanks{Corresponding author.} \and
Baojing Liu\inst{2}\orcidID{0009-0007-1444-7267}
}

\authorrunning{H. You and B. Liu}

\institute{
Graduate School of Arts and Sciences\\
Columbia University, New York, USA\\
\email{hy2854@columbia.edu}
\and
School of Artificial Intelligence\\
Hebei Institute of Communications, Shijiazhuang, PR China\\
\email{liubj@hebic.edu.cn}
}

\maketitle              
\begin{abstract}

Exemplar-free class-incremental learning enables models to learn new classes over time without storing data from old ones. As multimodal graph-structured data becomes increasingly prevalent, existing methods struggle with challenges like catastrophic forgetting, distribution bias, memory limits, and weak generalization. We propose \textbf{MCIGLE}, a novel framework that addresses these issues by extracting and aligning multimodal graph features and applying Concatenated Recursive Least Squares for effective knowledge retention. Through multi-channel processing, MCIGLE balances accuracy and memory preservation. Experiments on public datasets validate its effectiveness and generalizability.


\end{abstract}
\section{Introduction}

Class-Incremental Learning (CIL) addresses the challenge of incrementally learning new classes without forgetting previously learned ones, especially without access to historical data \cite{mittal2021essentials, zhou2024class}. It has broad applications, including dynamic classification, continual learning, and privacy-sensitive scenarios \cite{masana2022class}. Among CIL variants, exemplar-free learning is particularly challenging due to its strict constraint of not retaining any old class samples \cite{petit2024analysis}.

Without historical samples, exemplar-free CIL cannot rely on replay mechanisms and becomes more susceptible to catastrophic forgetting \cite{petit2023fetril}. Parameter updates for new classes may interfere with learned representations, degrading performance on old classes \cite{sun2023exemplar}. As a result, the model must rely on parameter regularization, knowledge distillation, or generated features-requiring strong generalization.

With the growth of large-scale multimodal data, CIL has increasingly focused on learning from diverse modalities (e.g., text, images, audio) while retaining prior knowledge \cite{d2023multimodal, pian2023audio}. In many real-world scenarios, such data form structured relationships-like user interactions or item co-occurrence—that are naturally represented as graphs \cite{wang2023semantic}. Graphs support dynamic topology, making them well-suited for continual learning. However, integrating graph-structured multimodal data remains underexplored. Existing approaches often rely on fixed GNN architectures, which lack scalability and adaptability, exacerbating forgetting in evolving tasks \cite{belouadah2019il2m, lei2020class}.

\begin{figure*}[htbp]
  \centering
  \includegraphics[width=\textwidth, height=0.55 \textwidth]{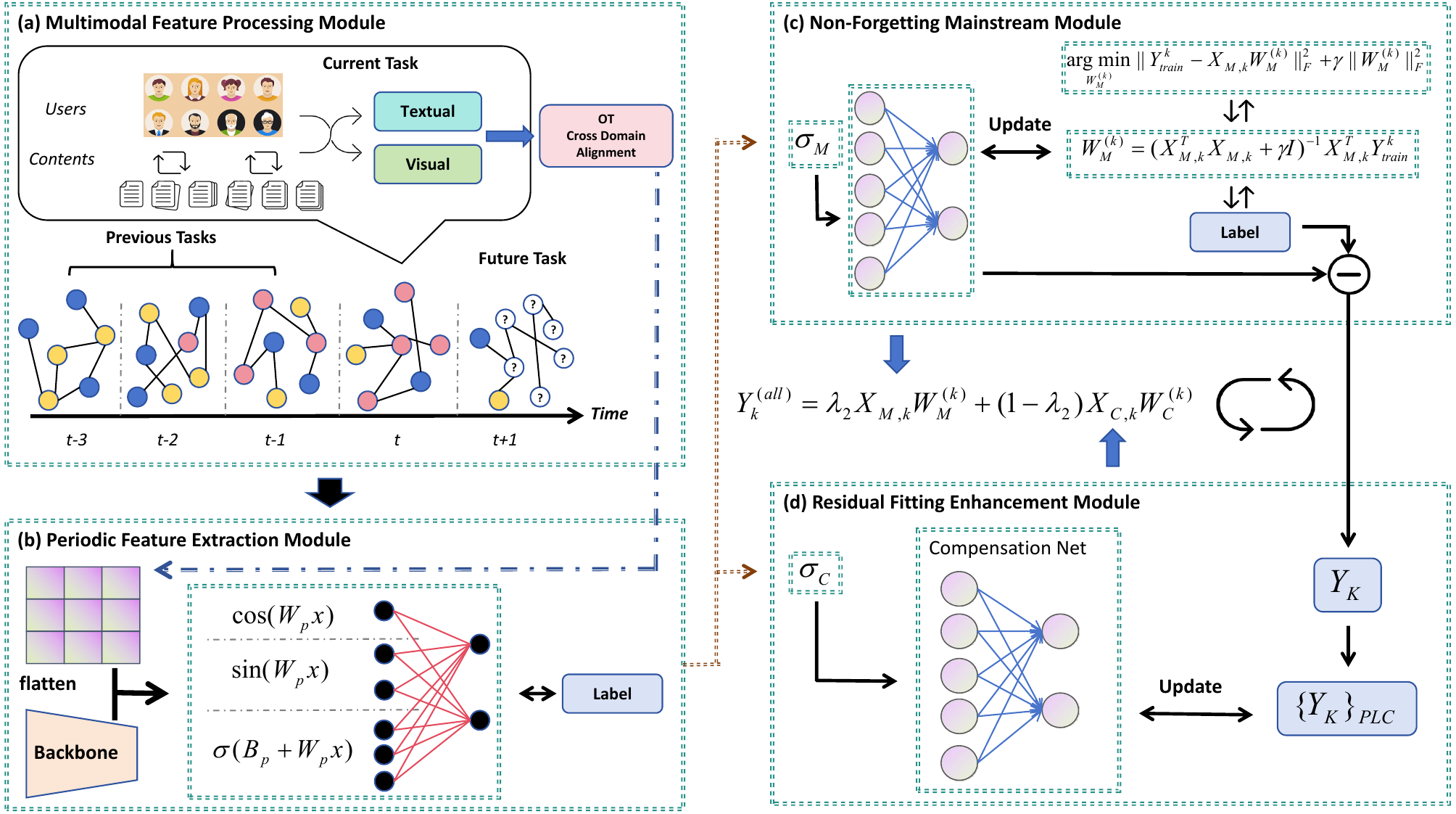}
  \caption{Framework of \textbf{MCIGLE}.}
  \label{Flowchart-1}
\end{figure*}

To address the existing challenges, we integrate multiple key modules in this paper and propose a novel framework. Our main contributions are given:

\begin{enumerate}[label=(\roman*)] 
    \item We propose a novel multimodal exemplar-free class-incremental graph learning framework, \textbf{MCIGLE}, illustrated in Figure~\ref{Flowchart-1}.
    \item To capture and align multimodal graph-structured features, we design a Multimodal Feature Processing Module and a Periodic Feature Extraction Module. Additionally, a Non-Forgetting Mainstream Module, based on Concatenated Recursive Least Squares, and a Residual Fitting Enhancement Module help mitigate forgetting and compensate for historical knowledge.
    \item Extensive comparative and ablation experiments on four public datasets validate the effectiveness, robustness, and generalization capability of MCIGLE.
\end{enumerate}

\section{Multimodal Class-Incremental Learning}

\subsection{Multimodal Feature Processing Module}

This module processes sequential multimodal graph structures with textual and visual modalities \cite{cai2022multimodal}, aligning information across modalities to support node classification in evolving graphs under continual learning.

Each modality $m \in \{v, t\}$ is represented as a graph $G_m=(V_m, E)$, where $V_m=\{X^m \mid X^m \in I_m\}$ and $E=\{(i, j)\}$. For the $l$-th layer, the feature update for node $u$ in modality $m$ is:

\begin{equation}
    h_{u, m}^{(l)}=\sigma\left(W_m^{(l)} \cdot \Phi_{l, m}\left(\left\{h_{v, m}^{(l-1)}: v \in N(u)\right\}\right)\right),
\end{equation}

Here, $\Phi_{l, m}$ aggregates neighbor features using correlation-based weighting:

\begin{equation}
    \Phi_{l, m}\left(\left\{h_{v, m}^{(l-1)} \right\}\right)=\operatorname{Agg}\left(\left\{e_{u, v, m}^{(l)} h_{v, m}^{(l-1)} \right\}\right),
\end{equation}

where $e_{u, v, m}^{(l)}$ is the correlation coefficient.

We align and fuse features via optimal transport. Specifically, visual features $h_{u, v}$ are projected to the textual feature space using the transport plan $P^{*}$:

\begin{equation}
    h_u^{\prime}=h_{u, v \rightarrow t} \| h_{u, t}, \quad \text{where } h_{u, v \rightarrow t} = P^{*} h_{u, v},
\end{equation}

and

\begin{equation} \label{dbot}
    P^{*} = \underset{\mathbf{P} \in \mathcal{P}}{\arg \min} \sum_{i,i',j,j'} \mathbf{P}_{ij} \left( \lambda_1 \mathbf{C}(x_i, y_j) + (1 - \lambda_1) \mathbf{L}(x_i, y_j, x_{i'}, y_{j'}) \mathbf{P}_{i'j'} \right) + \epsilon W(\mathbf{P}).
\end{equation}

where $\mathbf{C}$ is cost matrix, $\mathbf{L}$ the node similarity, and $W$ the regularization term.

Finally, node labels are predicted by:

\begin{equation}
    \hat{y}_u=\operatorname{softmax}\left(\tanh \left(W h_u^{(l)}+b\right)\right),
\end{equation}

with $W$ and $b$ being trainable parameters.

\subsection{Periodic Feature Extraction Module}

After obtaining node-level embeddings, we flatten them into a global feature matrix and normalize the input. The data is then passed through neural layers based on Fourier analysis \cite{dong2024fan}, with each layer defined as:

\begin{equation*}
    \phi^{(l)}(x)=\left[\cos \left(W_p^{(l)} x\right)\left\|\sin \left(W_p^{(l)} x\right)\right\| \sigma\left(B_{\bar{p}}^{(l)}+W_{\bar{p}}^{(l)} x\right)\right],
\end{equation*}

where $x$ is the input, $\sigma$ is a nonlinear activation (e.g., GELU or ReLU), and $W_p$, $W_{\bar{p}}$, $B_{\bar{p}}$ are trainable parameters. The full output is:

\begin{equation}
    X_{\text{FAN}} = \phi^{(L)} \circ \cdots \circ \phi^{(1)}(X_{\text{norm}}).
\end{equation}

We train the network using backpropagation with cross-entropy loss and optimizers such as AdamW. The output $X_{\text{FAN}}$ is passed to a classifier head for final predictions:

\begin{equation}
    \hat{Y} = \operatorname{softmax}(X_{\text{FAN}} \cdot W_{FCN}),
\end{equation}

with $W_{FCN}$ optimized jointly. After training, all parameters are frozen for later incremental phases.

Fourier-based layers effectively capture periodic patterns with fewer parameters than CNNs or MLPs, improving efficiency and representation quality. This enhances inputs for both the mainstream and compensation stages.

\subsection{Non-Forgetting Mainstream Module}

This module addresses class-incremental learning via analytical linear mapping using Concatenated Recursive Least Squares (C-RLS), enabling recursive weight updates without storing historical data and achieving performance comparable to joint training.

Let $X_{M,k}$ and $Y_{train}^k$ denote the input features and target labels at phase $k$, respectively. The optimization objective is:

\begin{equation}
    \underset{W_M^{(k)}}{\arg \min} \left\|Y_{train}^k - X_{M,k} W_M^{(k)}\right\|_F^2 + \gamma \left\|W_M^{(k)}\right\|_F^2,
\end{equation}

where $\gamma$ is a regularization parameter. The closed-form ridge regression solution is omitted for brevity.

To support efficient updates, we define an autocorrelation matrix $\Phi_k$ and a cross-correlation matrix $Z_k$, both updated recursively with forgetting factor $\beta$. The inverse $\Phi_k^{-1}$ is maintained using the Sherman–Morrison formula without direct matrix inversion \cite{gao2020recursive}.

The final recursive weight update is given by:

\begin{equation}
    \hat{W}_M^{(k)} = \hat{W}_M^{(k-1)} + k_k \left(Y_{train}^k - X_{M,k} \hat{W}_M^{(k-1)}\right),
\end{equation}

where the gain coefficient is:

\begin{equation}
    k_k = \frac{\Phi_{k-1}^{-1} X_{M,k}^T}{\beta + X_{M,k} \Phi_{k-1}^{-1} X_{M,k}^T}.
\end{equation}

\subsection{Residual Fitting Enhancement Module}

Due to the linear nature of the Non-Forgetting Mainstream Module, it may underfit complex samples \cite{xu2025drco}. To address this, we introduce a Residual Fitting Enhancement Module that captures the residual information missed by the main stream via independent feature projection and nonlinear transformations.

After training, the residual matrix is defined as:

\begin{equation}
    \widetilde{Y}_k = \left[0_{N_{0: k-1} \times d_{y, k-1}}, Y_k^{\mathrm{train}}\right] - X_{M,k} W_M^{(k)},
\end{equation}

representing prediction errors, where the zero matrix enforces phase-wise label exclusivity \cite{zhuang2024ds}.

The compensation stream generates embeddings via:

\begin{equation}
    X_{C,k} = \sigma_C\left(B\left(\operatorname{flat}\left(\operatorname{CNN}\left(X_k^{\text{train}}, W_{\mathrm{CNN}}\right)\right)\right)\right),
\end{equation}

with $\sigma_C$ (e.g., Tanh or Mish) different from the main stream’s activation to capture complementary features \cite{he20254s}.

To avoid error propagation from earlier phases, only the current-phase component of $\widetilde{Y}_k$ is retained:

\begin{equation}
    \left\{\widetilde{Y}_k\right\}_{\mathrm{PLC}} = \left[0_{N_{0:k-1} \times d_{y, k-1}}, \left(\widetilde{Y}_k\right)_{\mathrm{new}}\right].
\end{equation}

The compensation weights are updated using C-RLS:

\begin{equation*}
    W_C^{(k)} = W_C^{(k-1)'} + R_{C,k} X_{C,k}^T\left(\left\{\widetilde{Y}_k\right\}_{\mathrm{PLC}} - X_{C,k} W_C^{(k-1)'}\right),
\end{equation*}

where $R_{C,k}$ is the inverse correlation matrix of the compensation stream.

The final prediction combines both streams:

\begin{equation}
    \hat{Y}_k^{(\mathrm{all})} = \lambda_2 X_{M,k} W_M^{(k)} + (1 - \lambda_2) X_{C,k} W_C^{(k)},
\end{equation}

with $\lambda_2$ controlling the compensation contribution.

\section{Experiments}

\begin{figure}[htbp]
    \centering
    \begin{subfigure}[t]{0.4\linewidth} 
        \centering
        \includegraphics[width=\linewidth]{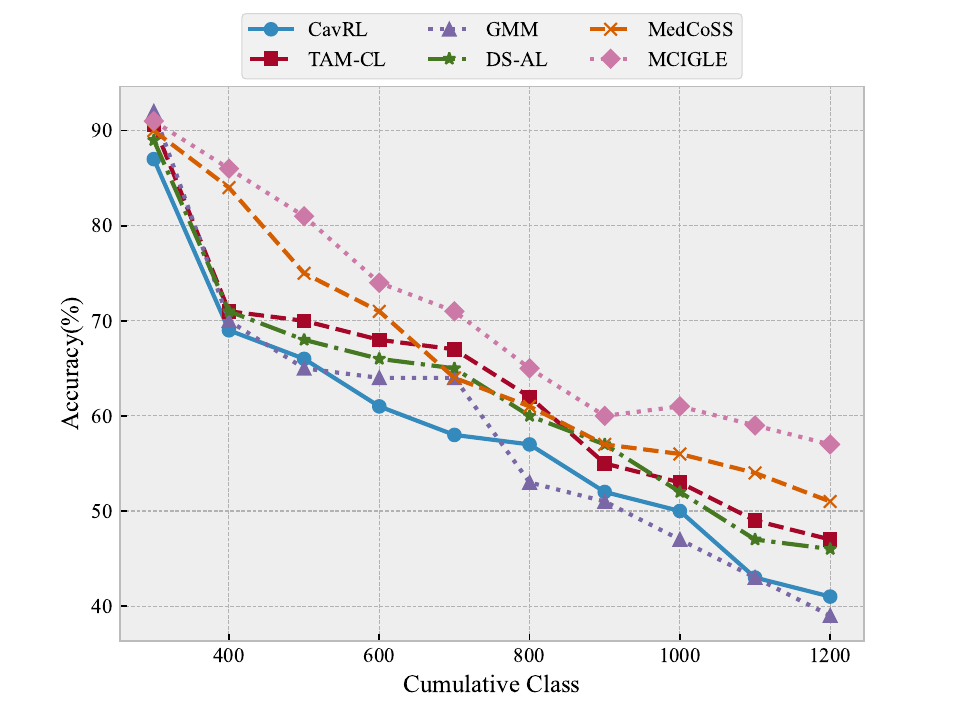}
        \caption{AudioSet-MI.}
        \label{AudioSet-MI}
    \end{subfigure}
    \hspace{0.05\linewidth}
    \begin{subfigure}[t]{0.4\linewidth}
        \centering
        \includegraphics[width=\linewidth]{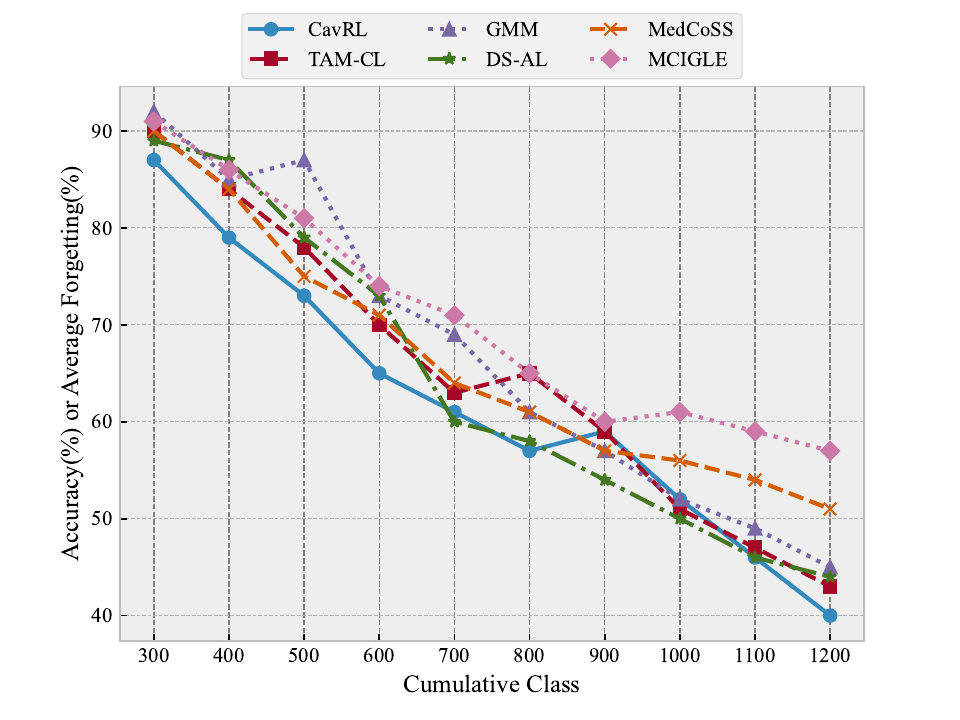}
        \caption{VoxCeleb.}
        \label{VoxCeleb}
    \end{subfigure}
    \caption{Changes in Prediction Accuracy of Various Models with Cumulative Class.}
    \label{fig:two_images}
\end{figure}

We evaluate \textbf{MCIGLE} on four public datasets-\textbf{COCO-QA}\footnote{\url{https://www.cs.toronto.edu/~mren/research/imageqa/data/cocoqa/}}, \textbf{VoxCeleb}\footnote{\url{https://www.robots.ox.ac.uk/~vgg/data/voxceleb/}}, \textbf{SNLI-VE}\footnote{\url{https://github.com/necla-ml/SNLI-VE}}, and \textbf{AudioSet-MI}\footnote{\url{https://research.google.com/audioset/index.html}}—and compare it with state-of-the-art methods including \textbf{CavRL} \cite{zhu2024learning}, \textbf{TAM-CL} \cite{cai2024dynamic}, \textbf{GMM} \cite{cao2024generative}, \textbf{DS-AL} \cite{zhuang2024ds}, and \textbf{MedCoSS} \cite{ye2024continual}.  As shown in Table~\ref{Overall Performance}, \textbf{MCIGLE} consistently outperforms baselines across most metrics and datasets, achieving high accuracy and low forgetting. The only exception is the Acc metric on AudioSet-MI, where CavRL performs better, which is expected given its specialization in audio-visual incremental learning. Figures~\ref{AudioSet-MI} and~\ref{VoxCeleb} further illustrate the accuracy trends over cumulative classes on AudioSet-MI and VoxCeleb.

\begin{table*}[htbp]
\centering
\caption{Overall Performance of our model \colorbox{LightPink}{MCIGLE} and the \colorbox{Peach}{baselines}.}
\label{Overall Performance}
\resizebox{\textwidth}{!}{
\begin{tabular}{l|cccc|cccc}
\toprule
\multirow{2}{*}{Method} & \multicolumn{4}{c}{COCO-QA} & \multicolumn{4}{c}{VoxCeleb} \\
\cmidrule(lr){2-5} \cmidrule(lr){6-9}
& Acc $\uparrow$ & F $\downarrow$ & BwF $\downarrow$ & $\mathbb{T}_{\mathbf{F}}(j \leftarrow i)$ $\downarrow$ & Acc $\uparrow$ & F $\downarrow$ & BwF $\downarrow$ & $\mathbb{T}_{\mathbf{F}}(j \leftarrow i)$ $\downarrow$ \\
\midrule
CavRL & 0.739 & 0.175 & \cellcolor{Peach}0.297 &  \cellcolor{Peach}0.250 & \cellcolor{Peach}0.915 & 0.237 & 0.249 & 0.273 \\
TAM-CL & 0.741 & 0.199 & 0.363 & 0.267 & 0.860 & \cellcolor{Peach}0.213 & 0.291 & 0.204 \\
GMM & 0.698 & \cellcolor{Peach}0.154 &  0.413 & 0.312 & 0.832 & 0.305 & 0.314 & 0.357 \\
DS-AL  & \cellcolor{Peach}0.749 & 0.269 & 0.321 & 0.288 & 0.904 & 0.246 & 0.285 & 0.190 \\
MedCoSS  & 0.721 & 0.188 & 0.306 & 0.274 & 0.892 & 0.317 & \cellcolor{Peach}0.247 & \cellcolor{Peach}0.185 \\
\textbf{MCIGLE} & \cellcolor{LightPink} \textbf{0.782} & \cellcolor{LightPink}\textbf{0.094} & \cellcolor{LightPink}\textbf{0.273} & \cellcolor{LightPink}\textbf{0.148} & \cellcolor{LightPink}\textbf{0.927} & \cellcolor{LightPink}\textbf{0.193} & \cellcolor{LightPink}\textbf{0.235} & \cellcolor{LightPink}\textbf{0.164} \\
\midrule\midrule
Method & \multicolumn{4}{c}{SNLI-VE} & \multicolumn{4}{c}{AudioSet-MI} \\
\midrule
CavRL & 0.740 & 0.258 & 0.302 &  0.204 & \cellcolor{Peach}0.685 & 0.194 & 0.455 & 0.211 \\
TAM-CL & 0.697 & 0.192 & \cellcolor{Peach}0.247 & 0.305 & 0.644 & \cellcolor{Peach}0.154 & 0.409 & 0.182 \\
GMM & 0.725 & 0.190 &  0.387 & \cellcolor{Peach}0.196 & 0.599 & 0.216 & \cellcolor{Peach}0.398 & 0.208 \\
DS-AL  & \cellcolor{Peach}0.752 & 0.183 & 0.395 & 0.251 & 0.637 & 0.165 & 0.493 & \cellcolor{Peach}0.167 \\
MedCoSS  & 0.641 & \cellcolor{Peach}0.140 & 0.372 & 0.209 & 0.671 & 0.251 & 0.413 & 0.253 \\
\textbf{MCIGLE} & \cellcolor{LightPink} \textbf{0.767} & \cellcolor{LightPink}\textbf{0.131} & \cellcolor{LightPink}\textbf{0.195} & \cellcolor{LightPink}\textbf{0.182} & \cellcolor{LightPink}\textbf{0.683} & \cellcolor{LightPink}\textbf{0.095} & \cellcolor{LightPink}\textbf{0.352} & \cellcolor{LightPink}\textbf{0.149} \\
\bottomrule
\end{tabular}
}
\end{table*}

\begin{figure}[htbp]
    \centering
    \begin{subfigure}[t]{0.48\linewidth}
        \centering
        \includegraphics[height=4.5cm]{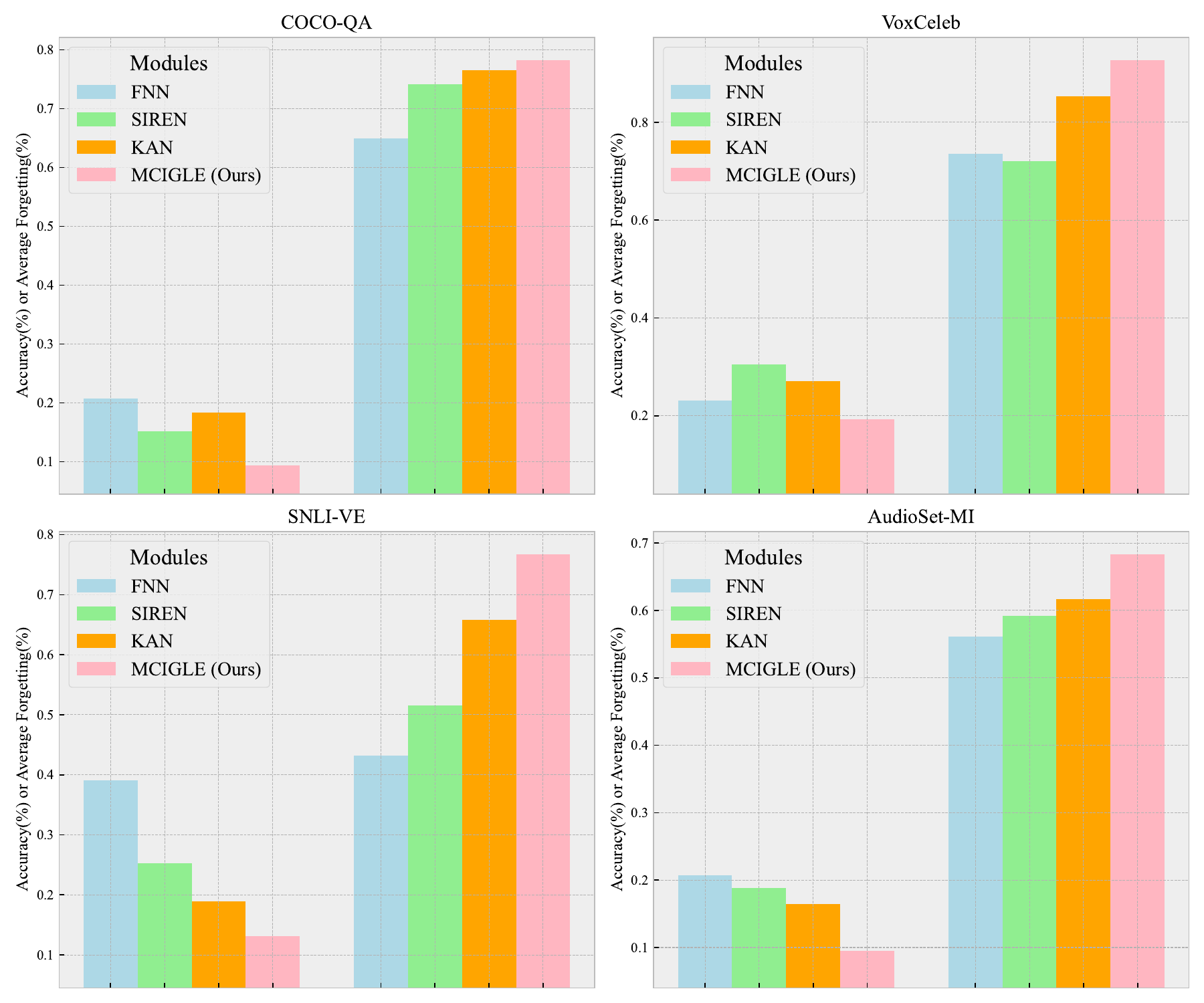}
        \caption{Ablation study replacing Fourier network with FNN, SIREN, and KAN.}
        \label{Ablation-1}
    \end{subfigure}
    \hspace{0\linewidth}
    \begin{subfigure}[t]{0.48\linewidth}
        \centering
        \includegraphics[height=4.5cm]{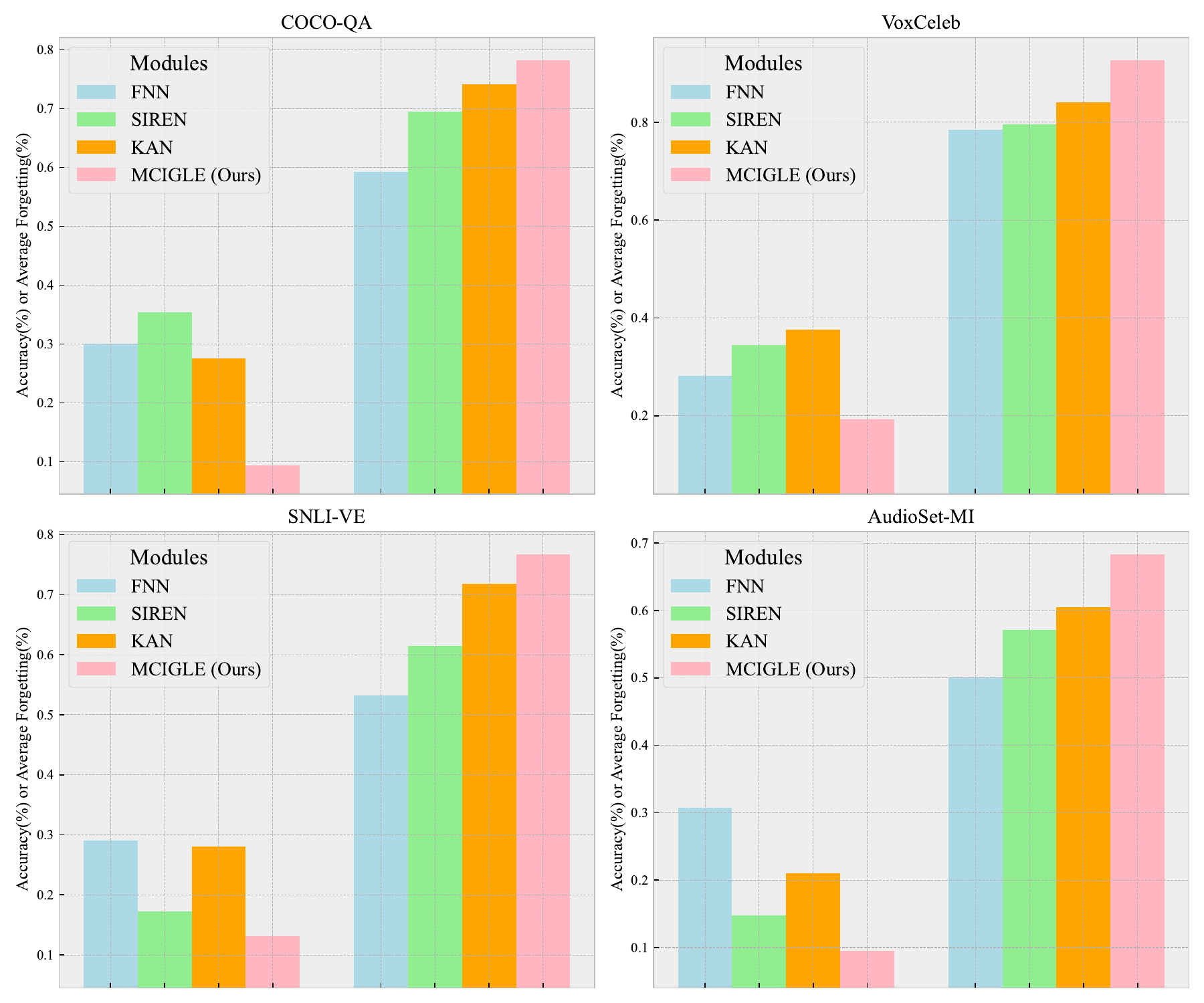}
        \caption{Ablation study replacing C-RLS with TA, UCL, and DGR.}
        \label{Ablation-2}
    \end{subfigure}
    \caption{Ablation results on accuracy and forgetting across model variants.}
    \label{fig:two_images_2}
\end{figure}

To verify component effectiveness, we conducted ablation studies by replacing key modules with representative alternatives. For Periodic Feature Extraction Module, we compared our Fourier-based network with \textbf{KAN} \cite{liu2024kan}, \textbf{FNN} \cite{silvescu1999fourier}, and \textbf{SIREN} \cite{russwurm2023geographic}, as shown in Figure~\ref{Ablation-1}. Results demonstrate superior accuracy and lower forgetting, validating our design for modeling temporal evolution.

We also replaced the C-RLS-based mechanism with classical forgetting mitigation approaches, including knowledge distillation \cite{szatkowski2024adapt}, parameter regularization \cite{ahn2019uncertainty}, and memory replay \cite{shin2017continual}. As shown in Figure~\ref{Ablation-2}, our model consistently achieves a lower forgetting rate, confirming the robustness of the C-RLS design.

\section{Conclusion}

This paper integrates multiple key modules to propose \textbf{MCIGLE}, an exemplar-free multimodal class-incremental graph learning framework. It addresses several critical challenges in continual learning and achieves promising results on public datasets, laying a solid foundation for broader applications in this field.

\bibliographystyle{splncs04}
\bibliography{mybibliography}

\end{document}